\documentclass[
]{ceurart}

\usepackage{listings}
\usepackage{graphicx}
\usepackage{subcaption}

\begin{document}

\copyrightyear{2025}
\copyrightclause{Copyright © 2025 for this paper by its authors. Use permitted under Creative Commons License Attribution 4.0 International (CC BY 4.0).}

\conference{MiGA@IJCAI25: International IJCAI Workshop on 3rd Human Behavior Analysis for Emotion Understanding, August 29, 2025, Guangzhou, China.}

\title{Multi-Track Multimodal Learning on iMiGUE: Micro-Gesture and Emotion Recognition}


\author[1]{Arman Martirosyan}[%
orcid=0009-0002-0969-585X,
email=martirosyan.arman@student.rau.am,
]
\cormark[1]

\author[1]{Shahane Tigranyan}[%
orcid=0000-0003-1536-9954,
email=shahane.tigranyan@rau.am,
]

\address[1]{Russian - Armenian University, Yerevan, Armenia}
\address[2]{ISP RAS, Moscow, Russia
}
\address[3]{AIRI, Moscow, Russia}
\address[4]{Sber AI Lab, Moscow, Russia}
\address[5]{HSE University, Moscow, Russia}
\address[6]{Innopolis University, Innopolis, Russia}
\address[7]{ISP RAS Research Center for Trusted Artificial Intelligence, Moscow, Russia}

\author[2]{Maria Razzhivina}[%
orcid=0000-0003-0188-6846,
email=mvrazzhivina@edu.hse.ru,
]

\author[5]{Artak Aslanyan}[%
orcid=0009-0004-8290-2557,
email={aaaslanyan_2@edu.hse.ru},
]

\author[6]{Nazgul Salikhova}[%
orcid=0009-0004-7246-280X,
email=n.salikhova@innopolis.university,
]

\author[2, 3]{Ilya Makarov}[%
orcid=0000-0002-3308-8825,
email=iamakarov@hse.ru,
]

\author[2, 4]{Andrey Savchenko}[%
orcid=0000-0001-6196-0564,
email=avsavchenko@hse.ru,
]

\author[2, 7]{Aram Avetisyan}[%
orcid=0000-0002-7066-6954,
email=a.a.avetisyan@ispras.ru,
]

\cortext[1]{Corresponding author.}

\begin{abstract}
  Micro-gesture recognition and behavior-based emotion prediction are both highly challenging tasks that require modeling subtle, fine-grained human behaviors, primarily leveraging video and skeletal pose data. In this work, we present two multimodal frameworks designed to tackle both problems on the iMiGUE dataset. For micro-gesture classification, we explore the complementary strengths of RGB and 3D pose-based representations to capture nuanced spatio-temporal patterns. To comprehensively represent gestures, video, and skeletal embeddings are extracted using MViTv2-S and 2s-AGCN, respectively. Then, they are integrated through a Cross-Modal Token Fusion module to combine spatial and pose information. For emotion recognition, our framework extends to behavior-based emotion prediction, a binary classification task identifying emotional states based on visual cues. We leverage facial and contextual embeddings extracted using SwinFace and MViTv2-S models and fuse them through an InterFusion module designed to capture emotional expressions and body gestures. Experiments conducted on the iMiGUE dataset, within the scope of the MiGA 2025 Challenge, demonstrate the robust performance and accuracy of our method in the behavior-based emotion prediction task, where our approach secured 2nd place.

\end{abstract}


\maketitle

\section{Introduction}

Micro-gesture recognition and behavior-based emotion prediction are both challenging and high-impact tasks that aim to interpret fine-grained human behaviors from visual data. These tasks are foundational for next-generation applications in human-computer interaction, affective computing, sign language interpretation, and immersive virtual or augmented reality environments. Despite their shared reliance on subtle behavioral cues, they differ in scope and modality requirements: micro-gesture recognition emphasizes the detection of low-amplitude movements in fingers, hands, or facial muscles using both RGB video and skeletal pose data, while behavior-based emotion prediction focuses solely on facial and contextual cues from video to infer emotional states in real-world settings, such as post-match interviews.

Unlike coarse gestures or full-body actions, micro-gestures involve nuanced and often ambiguous motions that demand high temporal and spatial resolution for reliable classification. Similarly, in emotion prediction, the ability to infer affective states from visual cues without relying on speech or textual data requires a fine-grained understanding of temporally distributed behaviors and subtle facial dynamics. In both tasks, conventional single-modality approaches fall short due to the complexity and ambiguity of the underlying signals.

Recent works\cite{liu2024prototype, chen2024fusion, chen2021crossvit} have demonstrated that combining multiple modalities, such as RGB video and pose or skeleton-based representations, can improve robustness and discrimination in gesture recognition. However, most existing fusion methods based on simple concatenation\cite{seddik2017hybrid}, late fusion\cite{liu2020latefusion}, or shallow attention\cite{shallowattention2021} fail to fully exploit the fine-grained complementary information between modalities, especially when the differences between gesture classes are subtle and context-dependent.

To address these challenges, we integrate a novel framework that combines multi-modal token-level feature learning with context-aware class refinement for precise micro-gesture recognition. We leverage MViTv2-S\cite{li2022mvitv2} and 2s-AGCN\cite{li2019agcn} encoders to extract temporally-aware visual tokens from RGB video frames and 3D skeletal pose sequences, capturing modality-specific dynamics in high resolution. To unify these heterogeneous modalities, we employ a Cross-Modal Token Fusion module that aligns and merges cross-modal tokens using multiple fusion heads based on spatial, semantic, and contextual relevance.

Furthermore, to refine class decision boundaries, we incorporate a Memory-Powered Refinement Module that learns to refine gesture classification based on accumulated knowledge of gesture representations.

We evaluate our approach on the iMiGUE\cite{liu2021imigue} dataset and observe strong performance, demonstrating its effectiveness for fine-grained micro-gesture classification. 

In addition to micro-gesture recognition, behavior-based emotion classification represents a complementary and equally challenging task in affective computing. Unlike conventional emotion recognition approaches that rely on facial expressions or vocal cues, behavior-based emotion prediction seeks to infer an individual’s emotional state based on nonverbal visual signals such as body posture and micro-gestures. Accurately modeling such subtle, temporally distributed behavioral patterns requires not only robust feature extraction from multiple visual streams but also effective fusion strategies that can capture inter-modal dependencies. In this work, we extend our framework to address this task using a dual-stream architecture that combines facial and contextual embeddings through iterative gated fusion. 

To evaluate the effectiveness of our approach in this setting, we conducted extensive experiments on the iMiGUE dataset, which provides a suitable benchmark for understanding behavior-driven emotions in real-world scenarios.

\section{Micro-gesture Classification}
\subsection{Task Definition}
The Micro-Gesture Classification task is a 32-way classification problem defined on short video clips containing fine-grained, subtle hand gestures. Given a video segment $V$ with $T$ frames, the goal is to predict a gesture label $y \in \{0, 1, \dots, 31\}$, where each class corresponds to a distinct micro-gesture, and class 99 indicates a \textit{non-gesture} (i.e., non-illustrative gesture). 

\subsection{Model Architecture}
\begin{figure}[htbp]
    \centering
    \begin{subfigure}{0.69\textwidth}
        \centering
        \includegraphics[width=\linewidth]{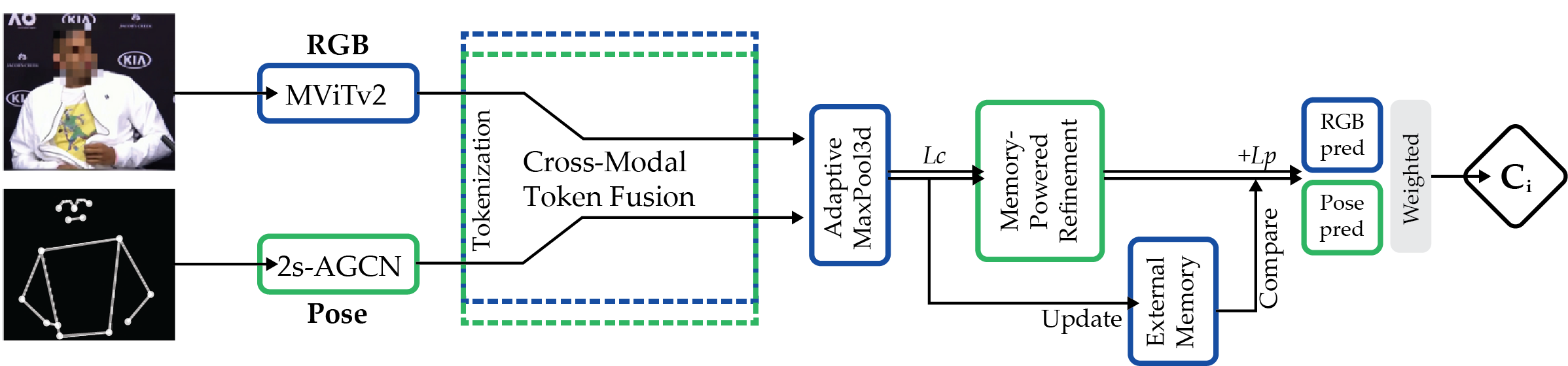}
        \caption{Overall architecture diagram.}
        \label{fig:dt1}
    \end{subfigure}
    \hfill
    \begin{subfigure}{0.29\textwidth}
        \centering
        \includegraphics[width=\linewidth]{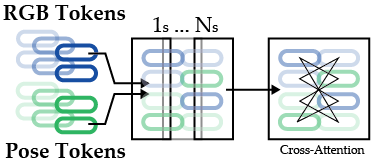}
        \caption{Cross-Modal token fusion module.}
        \label{fig:dt2}
    \end{subfigure}
    \caption{Detailed architecture diagrams for microgesture classification model.}
    \label{fig:comparison}
\end{figure}

We propose a novel multimodal framework (Figure~\ref{fig:dt1}) for fine-grained micro-gesture classification, which leverages both RGB and 3D skeletal pose representations. Our method builds upon recent advances in token-level fusion and multimodal refinement, integrating ideas from Multi-Criteria Token Fusion (MCTF)\cite{lee2024multi} and Context-Aware Prompt Learning (CAPL) \cite{lin2023contextaware} to improve alignment and discriminability of cross-modal features.

\subsection{Modality Encoding}
We extract modality-specific features using MViTv2-S and 2s-AGCN backbones. The RGB stream processes spatio-temporal clips of raw gesture videos, capturing fine-grained motion and appearance cues. The skeleton stream operates directly on 3D joint coordinates, using features extracted from a pretrained 2s-AGCN model.  

\subsection{Cross-Modal Token Fusion}
To effectively align and integrate the RGB and skeleton features, we apply the Cross-Modal Token Fusion module (CMTF) inspired by \cite{lee2024multi}. The module performs token-level cross-modal attention by considering multiple semantic and spatial criteria, dynamically attending to the most relevant tokens from the complementary modality. 

Given token sequences $\mathbf{T}_{\text{RGB}}$ and $\mathbf{T}_{\text{Pose}}$ extracted from the RGB and skeleton MViTv2 branches respectively, the Cross-Modal Token Fusion (CMTF) module produces a fused representation $\mathbf{T}_{\text{fused}}$ through dynamic token alignment:

$$
\mathbf{T}_{\text{fused}} = \text{CMTF}(\mathbf{T}_{\text{RGB}}, \mathbf{T}_{\text{Pose}})
$$
which is passed through a linear projection and temporal pooling to yield a compact feature vector for each gesture clip.

\subsection{Memory-Powered Refinement Module}

To further improve class separability, especially in fine-grained microgesture scenarios, we incorporate a Memory-Powered Refinement Module. This module maintains an external memory bank of prototypical embeddings per class. Inspired by prototype refinement approaches, it uses memory to refine predictions by comparing incoming features against stored high-confidence class exemplars.

    During the initial training epochs, the memory is populated with confident feature embeddings. In later epochs, for each input, we compare its features with the top-$k$ similar memory vectors (per predicted class) using cosine similarity. The comparisons are processed via multi-head self-attention to refine the features before classification. A refinement loss ($L_p$) is then computed, encouraging alignment with the most representative class features, and is combined with the standard classification loss ($L_c$):
\begin{equation}
\mathcal{L}_{\text{total}} = \mathcal{L}_{\text{c}} + \alpha \mathcal{L}_{\text{p}}
\label{eq:loss_t1_total}
\end{equation}
Here, $\mathcal{L}_{\text{c}}$ supervises classification based on averaged modality logits, while $\mathcal{L}_{\text{p}}$ encourages tighter intra-class clustering and increased inter-class margins in the feature space through both parametric (prototypes) and non-parametric (external memory) constraints.

To balance modality contributions in the final decision, we adopt a weighted late fusion strategy. While both RGB and pose classifiers output independent logits, we place higher emphasis on the pose branch due to its robustness in capturing subtle skeletal dynamics in microgestures. The final class prediction is computed as:
\begin{equation}
\mathbf{C}_i = w_{\text{pose}} \cdot \mathbf{y}_{\text{pose}} + w_{\text{RGB}} \cdot \mathbf{y}_{\text{RGB}}, \quad \text{where} \quad w_{\text{pose}} > w_{\text{RGB}}
\end{equation}
These weights are adjusted dynamically during training.

\begin{figure}[htbp]
    \centering
    \includegraphics[width=\textwidth]{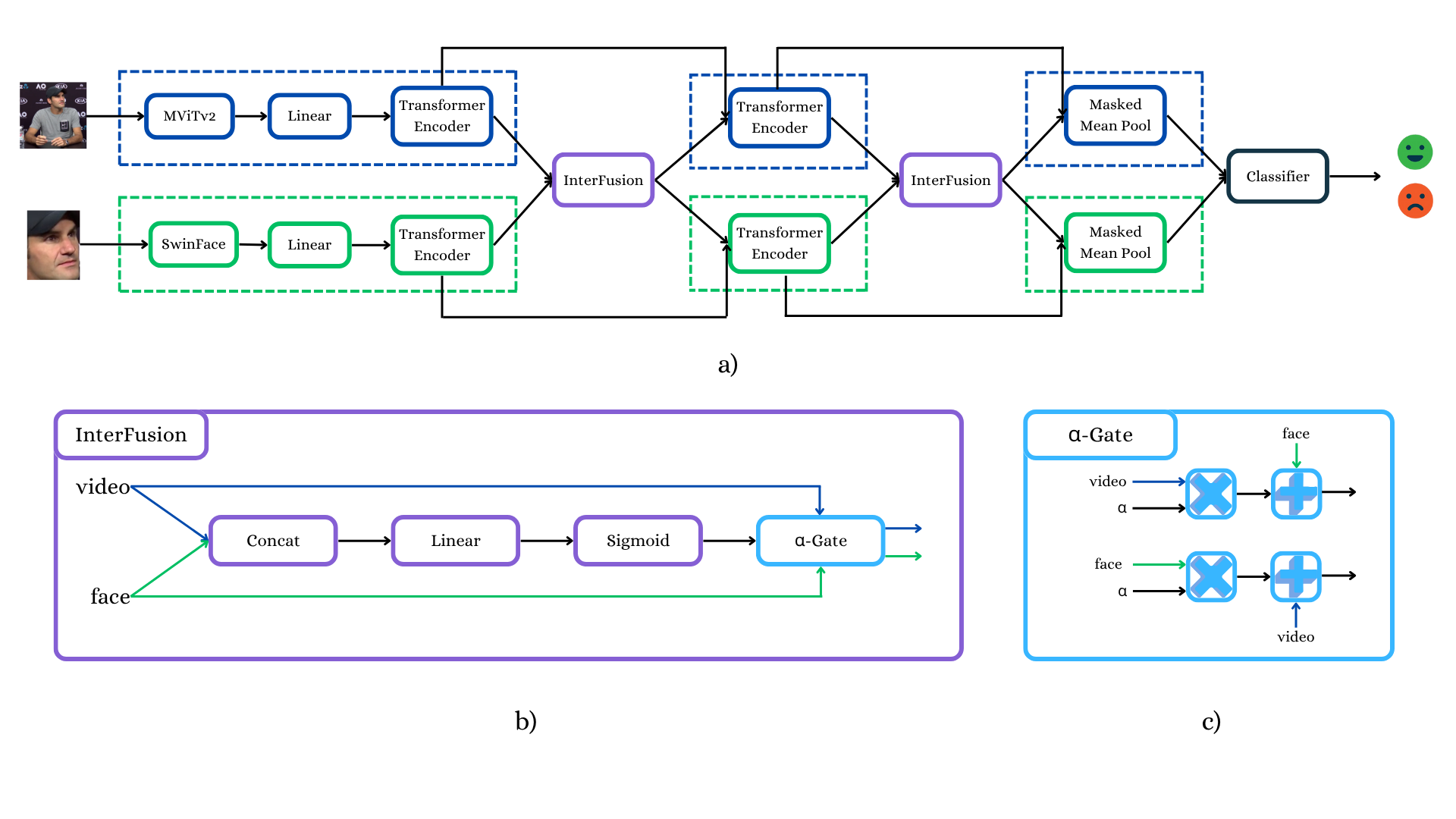}
    \caption{(a) Architecture of the proposed multimodal architecture for emotion recognition from video and facial features. (b) The structure of the InterFusion module. (c) The $\alpha$-Gate mechanism for information aggregation from two modalities.}
    \label{fig:architecture3}
\end{figure}

\section{Behavior-based Emotion Prediction}
\subsection{Task Definition}

The Behavior-Based Emotion Recognition task\footnote{https://www.kaggle.com/competitions/the-3rd-mi-ga-ijcai-challenge-track-3} is a binary classification problem defined on video sequences of post-match interviews. Given an interview video clip \( V \) containing \( T \) frames, the goal is to predict the match outcome label \( y \in \{0, 1\} \), where \( y = 1 \) indicates a win and \( y = 0 \) - a loss. Each video contains visible body and facial behaviors that may implicitly express the emotional state of the athlete.
The data is drawn from the iMiGUE dataset (details in Section), which captures fine-grained behavioral cues in press conferences of tennis players. The task is to develop a model that infers emotional signals relevant to the final outcome using only visual information, without access to audio or transcripts. 

\subsection{Model Architecture}
We propose a dual-stream transformer-based architecture for behavior-based emotion recognition that integrates contextual and facial information in parallel and integrates them through interfusion blocks. The model is designed to capture both intra-modal dynamics and inter-modal interactions through fusion mechanisms. The architecture is shown in Figure~\ref{fig:architecture3}(a).

The model takes as input two sequences:
\begin{itemize}
    \item \textbf{Contextual embeddings:} The video is processed using a pretrained MViTv2-S backbone, which operates on non-overlapping chunks of 16 frames and produces one embedding per chunk. This results in a sequence of frame-group representations \( \mathbf{X}^{\text{CTX}} \in \mathbb{R}^{T \times 768} \), where \( T \) denotes the number of 16-frame segments extracted from the video.

    \item \textbf{Facial embeddings :} Faces are detected in each frame using a pretrained YOLO-face model\footnote{https://github.com/akanametov/yolo-face}, and cropped face regions are encoded using SwinFace\cite{qin2023swinface}, a hierarchical vision transformer tailored for facial representation learning. While SwinFace produces one embedding per frame, we aggregate the features by averaging every 16 consecutive frame-level embeddings, resulting in a sequence \( \mathbf{X}^{\text{FACE}} \in \mathbb{R}^{T \times 512} \) that is temporally aligned with \( \mathbf{X}^{\text{CTX}}\). This ensures one embedding per 16-frame window for both modalities, enabling consistent cross-modal fusion.
\end{itemize}

Each stream begins with a linear projection followed by dropout for regularization. The projected embeddings are then passed through a modality-specific Transformer Encoder to get intra-modal temporal dependencies. 

To facilitate interaction between modalities, we employ the InterFusion module Figure~\ref{fig:architecture3}(b), a lightweight gated fusion block that enables bidirectional information exchange between the contextual and facial streams. Rather than using attention, InterFusion computes a shared element-wise gate from the concatenated inputs and performs a residual gated-sum fusion in both directions (shown in Figure~\ref{fig:architecture3}(c)). This allows each stream to selectively incorporate features from the other while preserving its own representation. This fusion step is followed by residual addition to preserve the original intra-modal information. The fused features are processed by an additional Transformer Encoder layer per stream, followed by a second InterFusion module and another residual connection. This iterative process allows deeper and progressively refined integration between the two modalities. 

After temporal modeling, a masked mean pooling operation is applied across time for both streams, producing fixed-size representations. These are concatenated and passed through a classification head to predict the class. 

This design enables the model to effectively capture subtle, temporally distributed cues from both body posture and facial expressions, which are critical in emotion-based outcome prediction.

\section{Experiments}

\section{Dataset: iMiGUE}
To evaluate our proposed methods, we utilized the iMiGUE (Identity-free Micro-Gesture Understanding and Emotion) dataset, introduced by Liu et al.\cite{liu2021imigue}. iMiGUE is a large-scale, identity-free video dataset specifically designed to facilitate research in micro-gesture recognition and emotion analysis. 

The dataset comprises 359 post-match press conference videos, featuring 72 athletes. Each clip captures spontaneous upper-body micro-gestures, such as "covering the face," "folding arms," or "crossing fingers," which are indicative of the athlete's emotional state following a win or loss. 

Annotations in iMiGUE encompass 32 behavioral classes, including 31 distinct micro-gesture categories and one non-micro-gesture category. In addition to gesture labeling, each sample is also annotated with a binary emotion label (positive or negative) based on contextual interpretation (e.g., winning or losing a match) and observable behavior. Each video frame is processed using the OpenPose toolkit to extract skeletal data. This skeletal representation includes two-dimensional spatial coordinates and confidence scores for each joint, facilitating detailed analysis of body movements. 

The dataset is split into predefined training, validation, and test sets to ensure consistent evaluation. Importantly, the test set is identity-independent and includes subjects that are not present in the training and validation sets. This split emphasizes generalization and mitigates overfitting to specific individuals, thus supporting the development of identity-agnostic models for micro-gesture and emotion recognition. For the emotion classification task, the test set comprises 100 videos, divided between 50 positive and 50 negative emotional states.

To address the challenge of imbalanced class distributions inherent in spontaneous behavior datasets, the authors propose an unsupervised learning method that aims to capture latent representations from the micro-gesture sequences themselves. This approach enhances the model's ability to generalize across diverse gesture categories and improves emotion recognition performance.

In the context of the MiGA 2025 Challenge, the organizers released an enhanced version of the iMiGUE dataset that includes unblurred facial regions. This version aims to facilitate comprehensive multimodal analysis by allowing researchers to incorporate facial expressions alongside body gestures in emotion recognition tasks. The availability of unblurred faces enables the development and evaluation of models that leverage both facial and gestural cues, potentially leading to more accurate and robust emotion understanding systems.

\subsection{Microgesture Classification Experiments}
Our experiments were conducted using a custom transformer-based classifier that integrates a Cross-Modal Token Fusion module and a refinement module with external memory. The model takes RGB and skeleton features from pre-trained MViTv2 and 2s-AGCN encoders and fuses them using multi-head self-attention to capture token-level cross-modal interactions. The fused representations are passed through adaptive pooling and modality-specific classifiers, with logits combined using a weighted sum for the final prediction.

To refine class boundaries, we introduce a prototypical memory module trained with a two-stage loss. During early epochs, the prototypical loss is disabled ($\alpha$=0.0) to allow memory buildup. Its influence is then gradually activated ($\alpha$=1.0) to enhance class separation via learned prototypes.

The final model configuration hyperparameters include: a hidden size of 512, 8 transformer heads, a memory size of 50 entries per class with top-5 nearest prototypes used for refinement, and a momentum of 0.9 for memory updates. Training was performed using the AdamW optimizer with a learning rate of 1e-4, weight decay of 1e-4, and ReduceLROnPlateau scheduler. 

To ensure stable training and better generalization, variable-length sequences were padded dynamically per batch; bucketing and balanced sampling were used to mitigate class imbalance.

\begin{table}[ht]
\captionsetup{font=small}
\centering

\begin{minipage}[t]{0.48\textwidth}
\scriptsize
\centering
\caption{Micro-gesture classification results on the iMiGUE.}
\begin{tabular}{lcc}
\hline
\textbf{Method} & \textbf{Modality} & \textbf{Top-1 Accuracy (\%)} \\
\hline
TSM~\cite{lin2019tsm} & RGB & 58.77 \\
VSwim-T~\cite{liu2022video} & RGB & 59.97 \\
VSwim-S~\cite{liu2022video} & RGB & 57.83 \\
VSwim-B~\cite{liu2022video} & RGB & 61.73 \\
\hline
ST-GCN~\cite{yan2018spatial} & Skeleton & 46.38 \\
ST-GCN++~\cite{duan2022pyskl} & Skeleton & 49.56 \\
StrongAug~\cite{duan2022pyskl} & Skeleton & 53.13 \\
AAGCN~\cite{shi2020skeleton} & Skeleton & 54.73 \\
CTR-GCN~\cite{chen2021channel} & Skeleton & 53.02 \\
DG-STGCN~\cite{duan2022dgstgcn} & Skeleton & 49.56 \\
PoseConv3D~\cite{duan2022revisiting} & Skeleton & 61.11 \\
\hline
DSCNet~\cite{cheng2024dense} & RGB \& Skeleton & 62.53 \\
Ours & RGB \& Skeleton & \textbf{62.87} \\
\hline
\end{tabular}
\end{minipage}
\hfill
\begin{minipage}[t]{0.48\textwidth}
\scriptsize
\centering
\caption{Fusion method and MR\textsuperscript{*} impact on accuracy.}
\begin{tabular}{lc}
\hline
\textbf{Fusion Method} & \textbf{Top-1 Accuracy (\%)} \\
\hline
Late Fusion & 58.81 \\
CMTF\textsuperscript{*} (w/o MR\textsuperscript{*}) & 62.23 \\
CMTF\textsuperscript{*} (w/ MR\textsuperscript{*}) & \textbf{62.87} \\
\hline
\end{tabular}
\vspace{0.5em}

{\scriptsize
CMTF\textsuperscript{*}: Cross-Modal Token Fusion;\\
MR\textsuperscript{*}: Memory Refinement Block.
}
\end{minipage}
\end{table}

\subsection{Emotion Prediction Experiments}

For the Track 3 task of the MiGA 2025 Challenge, the proposed multimodal fusion model was evaluated on the enhanced iMiGUE dataset, which includes unblurred facial regions. 

To enhance model performance, a comprehensive hyperparameter optimization was performed over the following search space: learning rate (1e-5 to 5e-4), transformer encoder depth (1 to 8), number of attention heads (2 to 8), dropout rate (0.1 to 0.5), and Focal Loss $\gamma$ parameter (0.5 to 1.5).
To address a severe class imbalance in the validation set, which initially contained only positive emotion samples, five negative samples were manually transferred from the training set to the validation set and excluded from training. This adjustment ensured more reliable validation performance across both classes. Class weights were calculated inversely proportional to the class frequencies in the adjusted training set.

Both Binary Cross-Entropy (BCE) and Focal Loss were evaluated, with Focal Loss demonstrating superior performance under the imbalanced setting.

The final model configuration used for comparison included a hidden size of 512, transformer encoder depth of 8, 4 attention heads, and a dropout rate of 0.5. The Focal Loss $\gamma$ parameter was set to 0.5. Training was conducted for up to 20 epochs with a batch size of 8, a learning rate of 1e-7, and early stopping with a patience of 7 epochs.

All experiments were conducted on a single NVIDIA A100 GPU with 80GB VRAM. Table~\ref{tab:track3_leaderboard} summarizes the comparison between our method, baseline approaches, and other submissions.

\begin{table}[h]
\centering
\caption{Leaderboard results from MiGA Challenge Track 3.}
\begin{tabular}{c l c}
\hline
\textbf{Rank} & \textbf{Team}         & \textbf{Score} \\
\hline
1            & backpacker            & 0.69230        \\
2            & \textbf{ISPCAST} (ours)        & 0.63461        \\
3            & haozhe bu             & 0.63461        \\
4            & gkdx2                 & 0.62500        \\
5            & KeXu2233              & 0.60576        \\
6            & Baseline              & 0.39423        \\
\hline
\end{tabular}
\label{tab:track3_leaderboard}
\end{table}

\section{Conclusion}
In this work, we presented two multimodal learning frameworks for addressing micro-gesture classification and behavior-based emotion recognition on the iMiGUE dataset. For the micro-gesture task, we proposed a novel architecture that fuses RGB and skeleton modalities via Cross-Modal Token Fusion and refines predictions through a memory-powered module leveraging class prototypes. Our method demonstrated strong performance, outperforming several prior RGB and skeleton-based baselines.

In the behavior-based emotion recognition task, we presented a dual-stream transformer-based model that jointly leverages facial and contextual visual information. The architecture incorporates iterative gated fusion via InterFusion modules to enable deep cross-modal interaction while preserving intra-modal information. The proposed model demonstrated competitive results in the MiGA 2025 Challenge Track, securing second place on the official leaderboard.

These results highlight the importance of fine-grained spatio-temporal modeling and multimodal interaction for subtle behavior understanding. 

\section*{Acknowledgment}

This work was supported by a grant, provided by the Ministry of Economic Development of the Russian Federation in accordance with the subsidy agreement (agreement identifier 000000C313925P4G0002) and the agreement with the Ivannikov Institute for System Programming of the Russian Academy of Sciences dated June 20, 2025 No. 139-15-2025-011.

\bibliography{references}

@inproceedings{lee2024multi,
  title     = {Multi-criteria Token Fusion with One-step-ahead Attention for Efficient Vision Transformers},
  author    = {Lee, Sanghyeok and Choi, Joonmyung and Kim, Hyunwoo J.},
  booktitle = {Conference on Computer Vision and Pattern Recognition},
  year      = {2024},
}

@inproceedings{shallowattention2021,
  title={Shallow Attention Mechanisms in Multimodal Learning},
  author={Tan, Y. and Wang, B.},
  booktitle={Proceedings of the 2021 International Conference on Multimodal Interaction},
  year={2021},
  pages={45--52}
}

@article{cheng2024dense,
  title     = {A dense-sparse complementary network for human action recognition based on RGB and skeleton modalities},
  author    = {Q. Cheng and J. Cheng and Z. Liu and Z. Ren and J. Liu},
  journal   = {Expert Systems with Applications},
  volume    = {244},
  pages     = {123061},
  year      = {2024},
  publisher = {Elsevier},
  doi       = {10.1016/j.eswa.2023.123061},
  url       = {https://www.sciencedirect.com/science/article/pii/S0957417423035637}
}

@article{liu2024prototype,
  title={Prototype Learning for Micro-gesture Classification},
  author={Liu, H. and Zhang, X. and Yu, M. and others},
  journal={arXiv preprint arXiv:2408.03097},
  year={2024}
}

@article{chen2024fusion,
  title={Multi-criteria Token Fusion with One-step-ahead Attention for Efficient Vision Transformers},
  author={Chen, Kai and Li, Xintao and Wang, Jianping},
  journal={arXiv preprint arXiv:2403.10030},
  year={2024}
}

@inproceedings{seddik2017hybrid,
  author = {Bassem Seddik and Sami Gazzah and Najoua Essoukri Ben Amara},
  title = {Hybrid Multi-modal Fusion for Human Action Recognition},
  booktitle = {ICIAR},
  year = {2017}
}

@inproceedings{chen2021crossvit,
  title={CrossViT: Cross-Attention Multi-Scale Vision Transformer for Image Classification},
  author={Chen, Chun-Fu and Xie, Qihang and Niu, Ming-Hsuan and others},
  booktitle={Proceedings of the IEEE/CVF International Conference on Computer Vision},
  pages={357--366},
  year={2021}
}

@inproceedings{liu2020latefusion,
  author={Liu, Yang and Chen, Qiang},
  title={Late Fusion for Robust Gesture Recognition},
  booktitle={ICPR},
  year={2020}
}

@inproceedings{li2022mvitv2,
  author={Li, Yanghao et al.},
  title={MViTv2: Improved Multiscale Vision Transformers for Classification and Detection},
  booktitle={CVPR},
  year={2022}
}

@inproceedings{lin2023contextaware,
  title={Context-Aware Prompt Learning for Vision-Language Models},
  author={Lin, Xiang and Zhou, Yixiao and Wang, Yue and Huang, Qian and Huang, Gao},
  booktitle={Proceedings of the IEEE/CVF International Conference on Computer Vision (ICCV)},
  year={2023}
}

@inproceedings{liu2021imigue,
  title={iMiGUE: An identity-free video dataset for micro-gesture understanding and emotion analysis},
  author={Liu, Xin and Shi, Henglin and Chen, Haoyu and Yu, Zitong and Li, Xiaobai and Zhao, Guoying},
  booktitle={Proceedings of the IEEE/CVF Conference on Computer Vision and Pattern Recognition (CVPR)},
  pages={13389--13398},
  year={2021}
}

@inproceedings{li2019agcn,
  title={Two-stream adaptive graph convolutional networks for skeleton-based action recognition},
  author={Li, Lei and Zhang, Xijie and Song, Yu and Chen, Yifan and Wang, Jiaying and Yang, Tao and Lee, Richie and Wang, Junbo and Huang, Junzhou},
  booktitle={Proceedings of the IEEE/CVF Conference on Computer Vision and Pattern Recognition (CVPR)},
  pages={12026--12035},
  year={2019}
}

@article{qin2023swinface,
  title={SwinFace: A multi-task transformer for face recognition, expression recognition, age estimation and attribute estimation},
  author={Qin, Lixiong and Wang, Mei and Deng, Chao and Wang, Ke and Chen, Xi and Hu, Jiani and Deng, Weihong},
  journal={IEEE Transactions on Circuits and Systems for Video Technology},
  volume={34},
  number={4},
  pages={2223--2234},
  year={2023},
  publisher={IEEE}
}

@inproceedings{lin2019tsm,
  title={TSM: Temporal shift module for efficient video understanding},
  author={Lin, Ji and Gan, Chuang and Han, Song},
  booktitle={Proceedings of the IEEE/CVF International Conference on Computer Vision},
  pages={7083--7093},
  year={2019}
}

@inproceedings{liu2022video,
  title={Video Swin Transformer},
  author={Liu, Ze and Ning, Jing and Cao, Yue and Wei, Yixuan and Zhang, Zheng and Lin, Stephen and Hu, Han},
  booktitle={Proceedings of the IEEE/CVF Conference on Computer Vision and Pattern Recognition},
  pages={3202--3211},
  year={2022}
}

@inproceedings{duan2022pyskl,
  title={Pyskl: Towards good practices for skeleton action recognition},
  author={Duan, Haodong and Wang, Jiaqi and Chen, Kai and Lin, Dahua},
  booktitle={Proceedings of the 30th ACM International Conference on Multimedia},
  pages={7351--7354},
  year={2022}
}

@article{duan2022dgstgcn,
  title={DG-STGCN: Dynamic spatial-temporal modeling for skeleton-based action recognition},
  author={Duan, Haodong and Wang, Jiaqi and Chen, Kai and Lin, Dahua},
  journal={arXiv preprint arXiv:2210.05895},
  year={2022}
}

@inproceedings{yan2018spatial,
  title={Spatial temporal graph convolutional networks for skeleton-based action recognition},
  author={Yan, Sijie and Xiong, Yuanjun and Lin, Dahua},
  booktitle={Proceedings of the AAAI conference on artificial intelligence},
  volume={32},
  number={1},
  year={2018}
}

@article{shi2020skeleton,
  title={Skeleton-based action recognition with multi-stream adaptive graph convolutional networks},
  author={Shi, Lei and Zhang, Yifan and Cheng, Jian and Lu, Hanqing},
  journal={IEEE Transactions on Image Processing},
  volume={29},
  pages={9532--9545},
  year={2020},
  publisher={IEEE}
}

@inproceedings{chen2021channel,
  title={Channel-wise topology refinement graph convolution for skeleton-based action recognition},
  author={Chen, Yuxin and Zhang, Ziqi and Yuan, Chunfeng and Li, Bing and Deng, Ying and Hu, Weiming},
  booktitle={Proceedings of the IEEE/CVF international conference on computer vision},
  pages={13359--13368},
  year={2021}
}

@inproceedings{duan2022revisiting,
  title={Revisiting skeleton-based action recognition},
  author={Duan, Haodong and Zhao, Yue and Chen, Kai and Lin, Dahua and Dai, Bo},
  booktitle={Proceedings of the IEEE/CVF conference on computer vision and pattern recognition},
  pages={2969--2978},
  year={2022}
}

\end{document}